\documentclass[sigconf, noacm]{acmart}
\AtBeginDocument{%
  }

\setcopyright{acmlicensed}
\copyrightyear{2026}
\acmYear{2026}
\acmDOI{XXXXXXX.XXXXXXX}
\acmConference[MM '26]{Proceedings of the 34th ACM International Conference on Multimedia}{November 10--14, 2026}{Rio de Janeiro, Brazil}
\acmBooktitle{Proceedings of the 34th ACM International Conference on Multimedia, November 10--14, 2026, Rio de Janeiro, Brazil}
\acmISBN{978-1-4503-XXXX-X/2018/06}

\usepackage{multirow}

\begin{document}

\title[OmniTrend: Content–Context Modeling for Scalable Social Popularity Prediction]{OmniTrend: Content–Context Modeling for Scalable Social Popularity Prediction}


\author{Liliang Ye}
\affiliation{%
  \institution{Huazhong University of Science and Technology}
  \city{Wuhan}
  \country{China}
}
\orcid{0009-0001-5076-2404}
\email{yll@hust.edu.cn}

\author{Guiyi Zeng}
\affiliation{%
 \institution{Huazhong University of Science and Technology}
 \city{Wuhan}
 \country{China}
 }
\orcid{0009-0001-6834-9829}
\email{tonyzengguiyi@hust.edu.cn}

\author{Yunyao Zhang}
\affiliation{%
  \institution{Huazhong University of Science and Technology}
  \city{Wuhan}
  \country{China}
}
\orcid{0009-0001-3412-9262}
\email{ikoyun@hust.edu.cn}

\author{Yi-Ping Phoebe Chen}
\affiliation{%
 \institution{La Trobe University}
 \city{Melbourne}
 \country{Australia}
 }
\orcid{0000-0002-4122-3767}
\email{phoebe.chen@latrobe.edu.au}

\author{Junqing Yu}
\affiliation{%
 \institution{Huazhong University of Science and Technology}
 \city{Wuhan}
 \country{China}
 }
\orcid{0000-0001-7057-0402}
\email{yjqing@hust.edu.cn}

\author{Zikai Song}
\authornote{Corresponding author.}
\affiliation{%
  \institution{Huazhong University of Science and Technology}
  \city{Wuhan}
  \country{China}
}
\orcid{0009-0006-6651-2027}
\email{skyesong@hust.edu.cn}

\renewcommand{\shortauthors}{First Author and Second Author}

\begin{abstract}
Predicting social media popularity requires understanding both the intrinsic appeal of content and the external context that determines how it is exposed to users. Existing methods focus on content signals but do not separate them from exposure-related patterns, which causes the learned representations to absorb platform-specific visibility effects and weakens both interpretability and cross-platform transfer. This paper introduces \textbf{OmniTrend}, a unified framework that models popularity as the joint outcome of content attractiveness and contextual exposure. The content module learns cross-modal representations from visual, audio, and textual cues to quantify intrinsic appeal, while the context module estimates exposure from exogenous signals such as posting time, author activity, topical trends, and retrieval-based neighborhood statistics. OmniTrend learns separate predictors for content attractiveness and contextual exposure and integrates them in the final popularity estimate, which makes the role of each factor explicit and supports robust transfer across image and video platforms. 
\end{abstract}

\ccsdesc[500]{Information systems~Multimedia information systems}
\ccsdesc[300]{Computing methodologies~Machine learning}
\ccsdesc[300]{Human-centered computing~Social media}
\ccsdesc[100]{Information systems~Data mining}

\keywords{Social Media Popularity Prediction, Multimodal Learning, Cross-platform Transfer}

\maketitle

\section{Introduction}
\label{sec:intro}

\begin{figure}[t]
    \centering
    \includegraphics[width=\columnwidth]{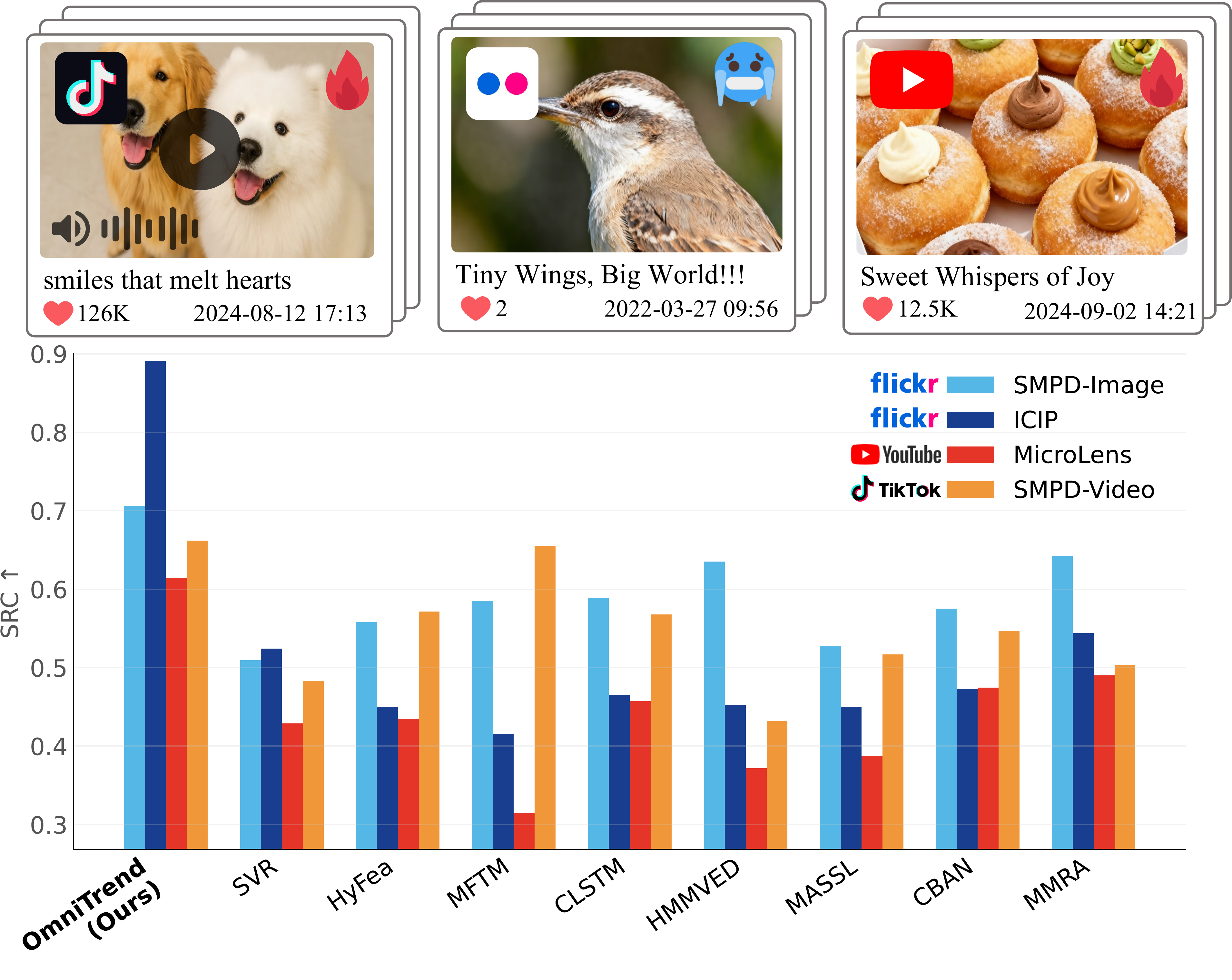}
    \caption{
        Examples of cross-platform social media posts and the Spearman rank correlation (SRC) of different methods on four datasets. OmniTrend achieves the highest rank correlation across both image and video benchmarks, demonstrating the effectiveness of joint content–context modeling.
    }
    \label{fig:teaser}
\end{figure}

Social media popularity prediction aims to estimate how audiences will engage with user-generated posts by analyzing their multimodal content~\cite{song3,song13,ReTrack,INTENT} and contextual signals~\cite{zhang2023improving,wu2024smpchallenge, song11, song14}.
Such prediction is essential for recommendation, content creation~\cite{yuan2026strucsum, yuan2025understanding, song10,HINT}, and trend analysis across platforms~\cite{yuan2025domainsum,song9, song8, song7, REFINE}, helping both creators and systems understand what drives audience engagement~\cite{xu2014forecasting,jing2017low,gelli2015image,zhang2023improving, song12,MELT}.

Despite steady progress, predicting social media engagement remains challenging because it depends on two closely coupled factors.
Platform algorithms decide how widely a post is shown, while users react to its content once it appears in their feed.
Observed engagement labels reflect both exposure and attractiveness. When exposure shifts across platforms or over time, the same content can map to different labels~\cite{ortis2019prediction,cheng2014can,haimovich2022time,xu2025smtpd}. Regression from content alone then learns unstable supervision.
Most multimodal models focus on learning content features~\cite{khosla2014makes,trzcinski2017predicting,inproceedings2018,xie2021micro,zhong2024predicting}, but they often do not explicitly separate platform-specific exposure effects from content signals, which limits interpretability and transferability across platforms~\cite{deng2023contentctr,tang2022knowledge,hsu2024revisiting,cheng2024retrieval,xu2025skapp}.
In contrast, exposure-aware models in recommender systems estimate visibility but lack multimodal understanding and are typically studied outside social media popularity prediction~\cite{khenissi2020modeling,krause2024mitigating,chen2023bias}.
Consequently, existing methods do not directly explain why similar content achieves very different levels of engagement across platforms or over time~\cite{hsu2024revisiting,ortis2019prediction,cheng2014can,tang2022knowledge,cheng2024casdo}, as illustrated in \autoref{fig:teaser}.

We revisit this challenge from the perspective of \emph{content–context modeling}.
From this viewpoint, social popularity arises from two complementary processes: how platforms determine exposure and how users perceive attractiveness.
Separating these processes clarifies the respective roles of external context and content itself, and provides a principled basis for scalable and interpretable popularity prediction in cross-platform settings.

Building on this idea, we propose \textbf{OmniTrend}, a unified framework that models content and context through distinct but complementary objectives.
The shared cross-platform content modeling module learns to represent the attractiveness of a post from visual and textual inputs, and is trained across platforms to improve cross-platform robustness.
The lightweight platform-specific context modeling module estimates exposure using external factors such as time, topic, and author activity, allowing efficient adaptation to new platforms through zero-shot or few-shot transfer.
The content module does not use metadata and remains portable across platforms.
To capture temporal and contextual patterns of popularity, the context module leverages relevant historical posts while keeping content information separate from exposure modeling.
Together, these designs enable OmniTrend to generate interpretable predictions and transferable representations for scalable social popularity prediction under platform shift.

Our contributions are summarized as follows:
\begin{enumerate}
    \item We propose a content–context framework that separates social popularity into exposure and attractiveness components, providing a clear view of how these two factors jointly determine engagement.
    \item We design a unified model that separately models content and context through a cross-platform content module and a platform-specific context module, enabling effective transfer and adaptation across different media environments.
    \item We introduce a retrieval-enhanced context modeling strategy that leverages relevant historical posts to capture temporal and contextual patterns of popularity while keeping content information separate from exposure modeling.
\end{enumerate}
Comprehensive experiments on multiple social media datasets demonstrate consistent improvements over existing multimodal and unified baselines, confirming that the proposed framework is effective, interpretable, and transferable.

\section{Related Works}
\label{sec:related}

\subsection{Social Media Popularity Prediction}

Recent work on social media popularity prediction has moved beyond simple multimodal fusion to incorporate temporal, social, and semantic factors that influence engagement~\cite{xie2021micro,tang2022knowledge,zhang2023improving,zhang2024contrastive,wu2024smpchallenge, song9}.
Attention-based architectures jointly encode intra-post content and inter-post dependencies to capture hierarchical relationships between items~\cite{zhang2023improving,tang2022knowledge}.
Other studies address inconsistencies between visual and textual modalities, adapting vision-language features to mitigate noise from weakly aligned pairs~\cite{hsu2024revisiting}.
Contrastive objectives have also been introduced to model implicit social influence, grouping posts by shared topical or author-related cues while separating unrelated ones~\cite{zhang2024contrastive}.
Temporal modeling has expanded from dynamic graph learning to benchmark design and continuous propagation modeling~\cite{ji2023community,jin2024predicting,xu2025smtpd,cheng2024casdo,haimovich2022time,jing2026mmcas}. This line of work shows that prediction quality depends not only on content but also on temporal alignment and changing exposure conditions.

Despite these advances, most existing methods treat content and context jointly, which entangles intrinsic appeal with exposure bias and reduces interpretability~\cite{deng2023contentctr,tang2022knowledge,hsu2024revisiting,chen2023bias}.
Our framework separates these factors through a content module that captures multimodal attractiveness and a context module that estimates exposure dynamics, complemented by retrieval-enhanced neighborhood aggregation to represent temporal diffusion.
This separation supports accurate and transferable prediction while offering a clearer interpretation of how social media popularity emerges.

\subsection{Multimodal Social Modeling}
Multimodal social modeling has evolved through several methodological stages.
\textit{\textbf{Early approaches}}~\cite{jing2017low,trzcinski2017predicting,chen2016micro,abousaleh2020multimodal}
relied on handcrafted visual and textual features combined through late fusion, which captured only coarse popularity trends and ignored cross-modal interactions.
\textit{\textbf{Deep feature methods}}~\cite{lai2020hyfea,ke2017lightgbm,inproceedings2018,zhang2022multi, long2026emomasemotionawaremultiagenthighstakes, song1,song2,song6,STABLE}
adopted convolutional and language encoders to extract richer multi-model representations~\cite{song4,song5,HABIT,OFFSET} but still processed each modality independently, limiting their ability to capture emotional or contextual factors that influence user engagement.
\textit{\textbf{Recent multimodal methods}}~\cite{xie2021micro,tang2022knowledge,deng2023contentctr,hsu2024revisiting,kayal2025large,wang2024research, song15}
used stronger joint encoders and fusion architectures for images, videos, and text, improving multimodal alignment and prediction quality.
They significantly improved cross-modal understanding but often mix content cues~\cite{Air-Know,ENCODER,HUD,TEMA} with platform-specific context, making it difficult to interpret or transfer popularity predictors across domains.
Recent extensions have explored adaptation of pretrained vision language features under weak image text alignment~\cite{hsu2024revisiting}, frame level multimodal modeling~\cite{deng2023contentctr}, prompt based prediction under temporal drift~\cite{xu2025smtpd,evopro2025sigir}, and broader benchmark settings that emphasize temporal consistency across platforms~\cite{haimovich2022time}.
Overall, these advances strengthen multimodal representation learning but still lack mechanisms to separate intrinsic content attractiveness from contextual exposure, which limits interpretation and cross-platform robustness.

\subsection{Retrieval-Augmented Popularity Modeling}
Retrieval-augmented methods extend popularity prediction by enriching each post with information from semantically related or temporally adjacent examples.
Earlier approaches also leveraged inter-post relations or historical dependencies~\cite{cheng2014can,chen2016micro,abousaleh2020multimodal}, where visual, textual, or temporal similarity helped connect related items.
While these structures helped capture diffusion patterns, they required dense user histories and were difficult to update as new posts appeared.
Recent work replaces static graphs with retrieval modules that dynamically search for relevant content from historical embeddings or memory banks~\cite{zhong2024predicting,cheng2024retrieval,xu2025skapp,xu2026jrpp}.
MMRA~\cite{zhong2024predicting} retrieves multimodal neighbors using joint image text similarity to enhance post representations, while RAGTrans~\cite{cheng2024retrieval} adds retrieved multimodal relations to popularity prediction.
Selective retrieval methods further improve context quality by filtering retrieved evidence, and recent joint optimization of retrieval and prediction reduces mismatch between retrieval objectives and final popularity estimation~\cite{xu2025skapp,xu2026jrpp}.
These efforts highlight the importance of leveraging relational context without explicit graph construction.
Our retrieval-enhanced design follows this principle by integrating neighborhood statistics from similar and earlier posts, allowing the model to represent temporal diffusion and social contagion more flexibly than static relational graphs.

\section{Challenge and Motivation}
\label{sec:challenge_motivation}

\paragraph{Challenge: popularity labels mix content and context.}
Social media popularity prediction aims to estimate the observed engagement $y_i$ of post $i$ from its multimodal signals and posting conditions~\cite{khosla2014makes,gelli2015image,xie2021micro,wu2024smpchallenge}. The central difficulty is that the label does not measure content quality alone. A post receives engagement only after it is exposed, then users react to what they see. Let $A_i$ denote the intrinsic attractiveness of the content and let $E_i$ denote its exposure opportunity. The observed popularity can then be viewed as the joint outcome of these two factors,
\begin{equation}
1 + y_i \propto A_i E_i.
\end{equation}
This relation explains why the same content may lead to very different engagement across platforms, time periods, or audiences~\cite{ortis2019prediction,cheng2014can,hsu2024revisiting,haimovich2022time,xu2025smtpd}. It also shows that the supervision signal used in popularity prediction is inherently mixed.

\paragraph{Challenge: content and context follow different mechanisms.}
From the perspective of content and context modeling, content and context play different roles in this process. Content determines how users respond after a post is shown, while context determines whether and how widely the post is shown. Let $c_i$ denote the multimodal content of post $i$ and let $x_i$ denote its contextual conditions, such as posting time, author activity, topical environment, and platform-specific delivery factors. A natural factorization is
\begin{equation}
\mathbb{E}[y_i \mid c_i, x_i] = g\big(f_c(c_i), f_x(x_i)\big),
\end{equation}
where $f_c(c_i)$ captures content attractiveness and $f_x(x_i)$ captures contextual exposure. These two terms are coupled in the final label, but they arise from different mechanisms. Content attractiveness reflects semantic appeal, which is relatively stable across platforms. Contextual exposure depends on external conditions whose distributions often change across platforms and over time.

\paragraph{Motivation: we separate content attractiveness from contextual exposure.}
This distinction motivates explicit separation. If a model directly regresses engagement from content alone, the learned representation must absorb not only intrinsic appeal but also platform dependent exposure bias~\cite{chen2023bias,khenissi2020modeling,krause2024mitigating}. Such entanglement weakens interpretability and reduces transferability. We therefore transform the count scale into log space and write
\begin{equation}
\tilde{y}_i = \log(1+y_i) = \alpha_i + \phi_i,
\end{equation}
where $\alpha_i$ denotes content attractiveness and $\phi_i$ denotes contextual exposure. This additive form follows the multiplicative relation above and provides a principled basis for separate estimation. At the expectation level, the same view becomes
\begin{equation}
\mathbb{E}[\tilde{y}_i \mid c_i, x_i] = f_{\alpha}(c_i) + f_{\phi}(x_i).
\end{equation}
Under this formulation, the content term can focus on platform robust semantic cues, while the context term captures external visibility patterns that vary with platform mechanisms and temporal conditions~\cite{cheng2024casdo,jing2026mmcas}. This separation gives a clearer interpretation of the prediction process and supports cross-platform transfer. Based on this motivation, we design the framework with a content module for $\alpha$, a context module for $\phi$, and a retrieval based neighborhood component that enriches exposure modeling without mixing raw content signals into the exposure estimator~\cite{zhong2024predicting,cheng2024retrieval,xu2025skapp,xu2026jrpp}.

\section{Method}
\subsection{Overview}

\begin{figure}[t!]
    \centering
    \includegraphics[width=\linewidth]{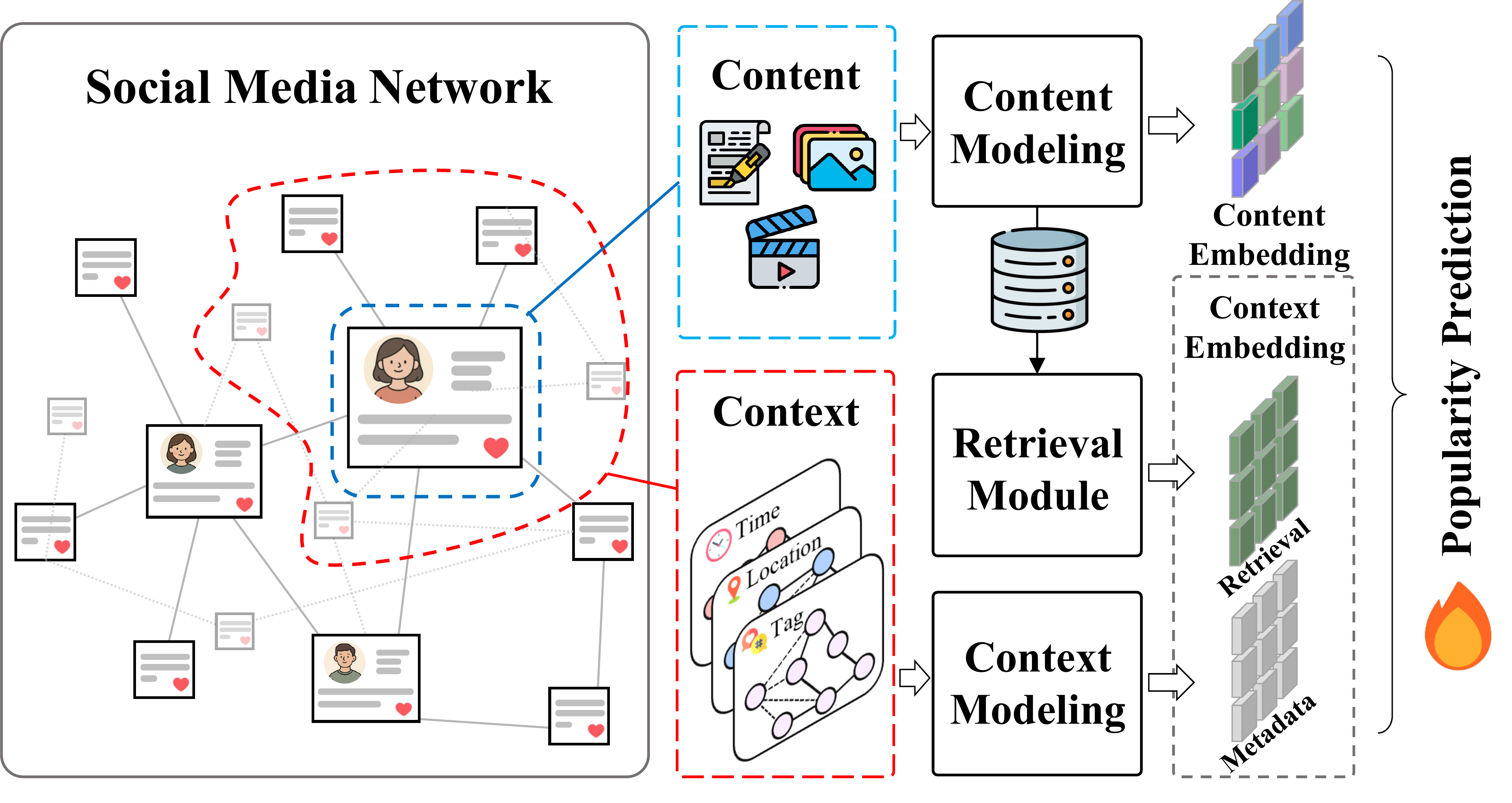}
    \caption{Overall architecture of the proposed content–context framework for social media popularity prediction.
  The \textbf{content} module models intrinsic attractiveness from multimodal inputs shared across platforms, while the \textbf{context} module estimates exposure from exogenous and platform-specific factors.
  A retrieval-augmented neighborhood module enriches the context representation with temporal cues without leaking content information.}
    \label{fig:pipeline}
\end{figure}

The challenge described in \autoref{sec:challenge_motivation} motivates a decomposition of observed popularity into content attractiveness and contextual exposure. Following that formulation, we estimate these two factors with separate modules and combine them to recover the final prediction. For post $i$, we write
\begin{equation}
\log (1 + y_i) = \alpha_i + \phi_i ,
\end{equation}
where $y_i$ denotes the observed engagement count, $\alpha_i$ captures the appeal derived from multimodal content, and $\phi_i$ represents exposure determined by external and platform-specific factors. The predicted engagement is recovered as
$\hat{y}_i = \exp(\hat{\alpha}_i + \hat{\phi}_i) - 1$
by combining the contributions from both modules.

The overall architecture, shown in \autoref{fig:pipeline}, contains three main components:
(1) a cross-platform content module to learn multimodal attractiveness shared across platforms,
(2) a platform-aware context module to estimate exposure from external signals,
and (3) a retrieval-enhanced context modeling module that incorporates temporal diffusion cues without leaking content features~\cite{zhong2024predicting,cheng2024retrieval}.

\subsection{Content Module}
\label{sec:alpha}
User engagement depends on how visual, auditory, and textual cues interact to influence user perception.
However, the composition of available modalities varies across platforms, as some are image-centric while others are video-centric.
Therefore, a unified attractiveness model should learn cross-modal interactions while remaining robust to missing modalities.

\begin{figure}[h]
\centering
\includegraphics[width=\linewidth]{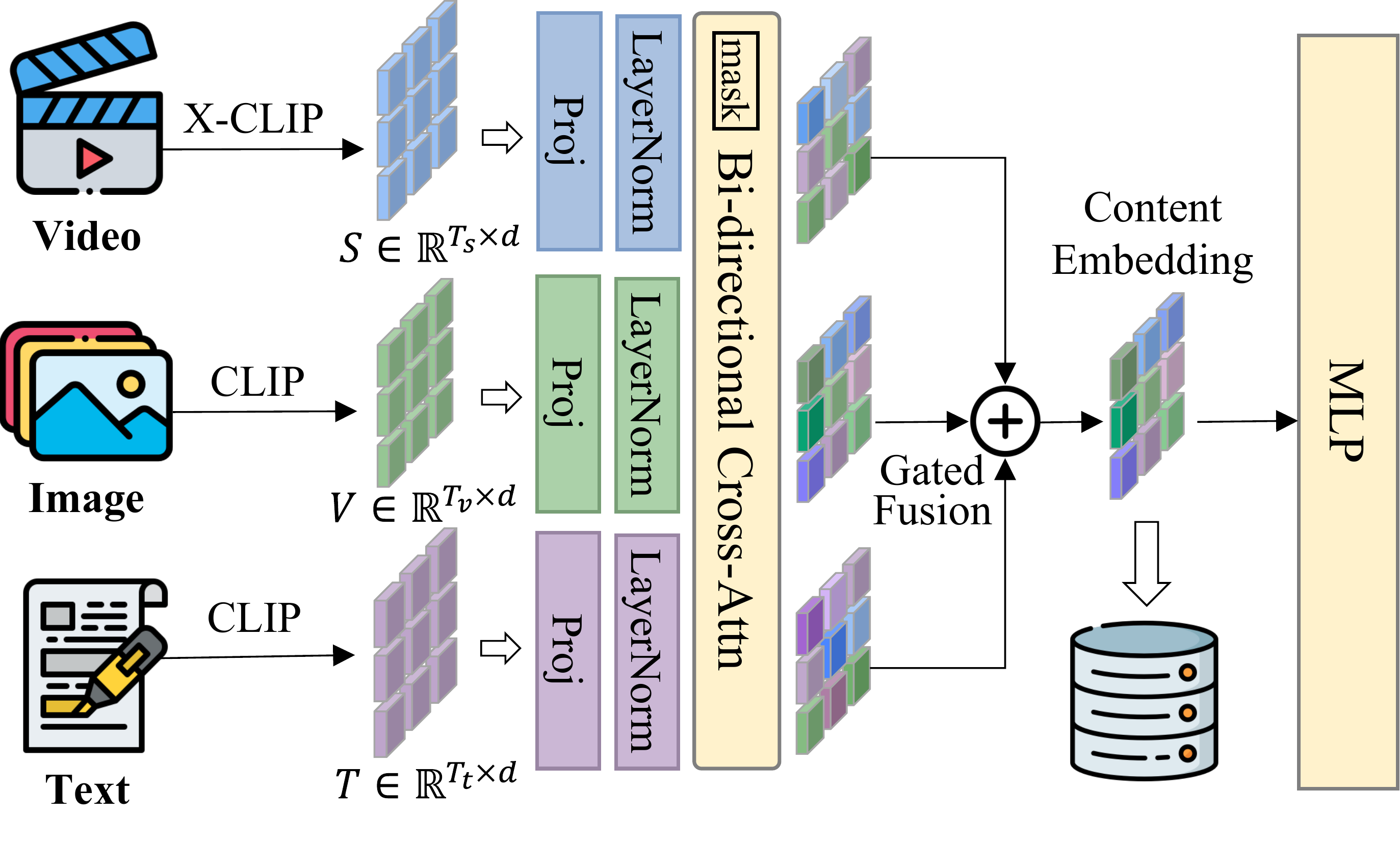}
\caption{
Architecture of the cross-platform content modeling module, which learns intrinsic attractiveness from multimodal inputs with cross-modal attention and a gated fusion block.
}
\label{fig:content}
\end{figure}

\autoref{fig:content} illustrates how the content module integrates multimodal cues into a unified attractiveness representation.
Visual, textual, and video features are first aligned through bidirectional cross-attention, then fused by a gated module that controls each modality’s contribution before predicting the attractiveness score.

\subsubsection{Input Encoders \& Normalization}
\label{sec:input_encoder}
We encode images, videos, and text using pretrained encoders and map all features to a shared embedding space of dimension $d=512$ to ensure cross-modal consistency.
Image features are extracted with CLIP~\cite{radford2021learning} and $\ell_2$-normalized so that $\lVert f_{\mathrm{img}} \rVert_2 = 1$.
Text features are obtained with BGE-M3 from structured prompts including title, category tags, location, and author biography fields.
Video features are derived from uniformly sampled frames encoded with X-CLIP~\cite{ma2022x} and averaged into a single video-level representation.
These representations are normalized to comparable scales across modalities, reducing domain differences at the input level.

\subsubsection{Cross-Modal Attention}
\label{sec:crossattn}
Let visual tokens be $V \in \mathbb{R}^{T_v \times d}$, textual tokens be $T \in \mathbb{R}^{T_t \times d}$ and video tokens be $S \in \mathbb{R}^{T_s \times d}$.
We apply bidirectional cross-attention to enable multimodal alignment and feature refinement:
\begin{equation}
\begin{aligned}
    \tilde{V} &= \mathrm{CrossAttn}(Q=V, K=T, V=T), \\
    \tilde{T} &= \mathrm{CrossAttn}(Q=T, K=V, V=V).
\end{aligned}
\end{equation}
For videos, the same operation is applied between $S$ and $T$, yielding $\tilde{S}$.

To handle platform-dependent modality availability, we define presence indicators $\mathbb{I}_{\mathrm{img}}, \mathbb{I}_{\mathrm{vid}}, \mathbb{I}_{\mathrm{txt}} \in \{0,1\}$.
Attention updates are applied only to available modalities.
Missing modalities are masked and removed from attention, fusion, and alignment losses to prevent noise propagation.
This masking strategy ensures stable training when certain modalities are absent.
It further reduces the risk that one modality dominates the attention updates when the others are sparse or noisy.

\subsubsection{Fusion \& Prediction}
We aggregate attended features from all available modalities into a unified representation.
Let $\mathbb{I}_m \in \{0,1\}$ indicate whether modality $m$ is available,
and define the fused token vector as
$u = [\mathbb{I}_{\mathrm{img}}\tilde{V}_{\mathrm{CLS}};\,\mathbb{I}_{\mathrm{txt}}\tilde{T}_{\mathrm{CLS}};\,\mathbb{I}_{\mathrm{vid}}\tilde{S}]$.
A gated fusion module combines them through
\begin{equation}
\label{eq:gatedfusion}
g = \sigma(W_g u + b_g), \quad
z = g \odot (W_z u + b_z),
\end{equation}
where $\sigma$ denotes the sigmoid gate, $\odot$ element-wise multiplication, and $z \in \mathbb{R}^{d}$ is the fused representation.
A lightweight $\mathrm{MLP}(\cdot)$ with GELU activation then maps $z$ to the attractiveness score $\hat{\alpha}$.

\subsubsection{Training Objectives}
Training is conducted jointly across multiple platforms with standardized labels to learn a platform-agnostic notion of appeal.
We combine complementary objectives to ensure stability, ranking fidelity, and cross-modal consistency.

\paragraph{Weighted Huber Loss.}
A weighted Huber loss~\cite{huber1992robust} mitigates label imbalance and emphasizes long-tail examples:
\begin{equation}
\label{eq:huber}
    L_{\mathrm{Huber}} =
        \begin{cases}
            \tfrac{1}{2} w_i r_i^2, & |r_i| \le \delta, \\
            w_i (\delta |r_i| - \tfrac{1}{2}\delta^2), & \text{otherwise},
        \end{cases}
\end{equation}
where $r_i = \hat{\alpha}_i - \tilde{y}_i$ denotes the residual between the predicted and normalized popularity labels, and $w_i$ is the per-sample weight based on label bins.

\paragraph{Pairwise Ranking Loss.}
To preserve popularity ordering, we apply a margin-based pairwise objective:
\begin{equation}
\label{eq:loss_pair}
L_{\mathrm{pair}} = \mathbb{E}_{(i,j)} \big[ \max(0, \gamma - (\hat{\alpha}_i - \hat{\alpha}_j)) \big],
\end{equation}
where $(i,j)$ are sampled within each batch from higher- and lower-popularity bins.

\paragraph{Cross-Modal Alignment.}
For samples with multiple modalities, we encourage feature consistency through a symmetric InfoNCE~\cite{oord2018representation} alignment:
\begin{equation}
L_{\mathrm{align}} =
\tfrac{1}{2}\!\left(
\ell_{\mathrm{NCE}}(h_{\mathrm{vis}}, h_{\mathrm{txt}}; \tau)
+ \ell_{\mathrm{NCE}}(h_{\mathrm{txt}}, h_{\mathrm{vis}}; \tau)
\right),
\end{equation}
where $h_m = \phi_m(z_m)/\|\phi_m(z_m)\|_2$ and $\phi_m$ is a modality-specific projection.
This encourages multimodal representations to remain semantically aligned even when trained across heterogeneous platforms.

\paragraph{Mean-Centering Regularization.}
To reduce domain-level bias and stabilize prediction scale, we apply a mean-centering term
$L_{\mu} = \left| \frac{1}{B} \sum_{i=1}^{B} \hat{\alpha}_i \right|$,
which penalizes shifts in the predicted mean attractiveness within each batch.

The total objective is
\begin{equation}
\label{eq:loss_alpha}
L_{\alpha} =
\lambda_{\mathrm{Huber}} L_{\mathrm{Huber}} +
\lambda_{\mathrm{pair}} L_{\mathrm{pair}} +
\lambda_{\mathrm{align}} L_{\mathrm{align}} +
\lambda_{\mu} L_{\mu}.
\end{equation}
The entire module is optimized end to end with the combined objective.

This design captures cross-modal interactions and remains effective under missing modalities.
Because the content module estimates intrinsic attractiveness that is independent of any platform-specific factors,
it can be directly transferred to unseen platforms in a zero-shot manner or refined with limited labeled data in a few-shot setting.
Further evidence appears in the supplementary material. A cross platform embedding visualization shows that image and video samples occupy an overlapping region in the learned content space. The supplementary training curves also show smooth convergence within a few epochs.

\subsection{Context Module: Platform-Aware Exposure}
\label{sec:phi}

\begin{figure}[!t]
\centering
\includegraphics[width=\linewidth]{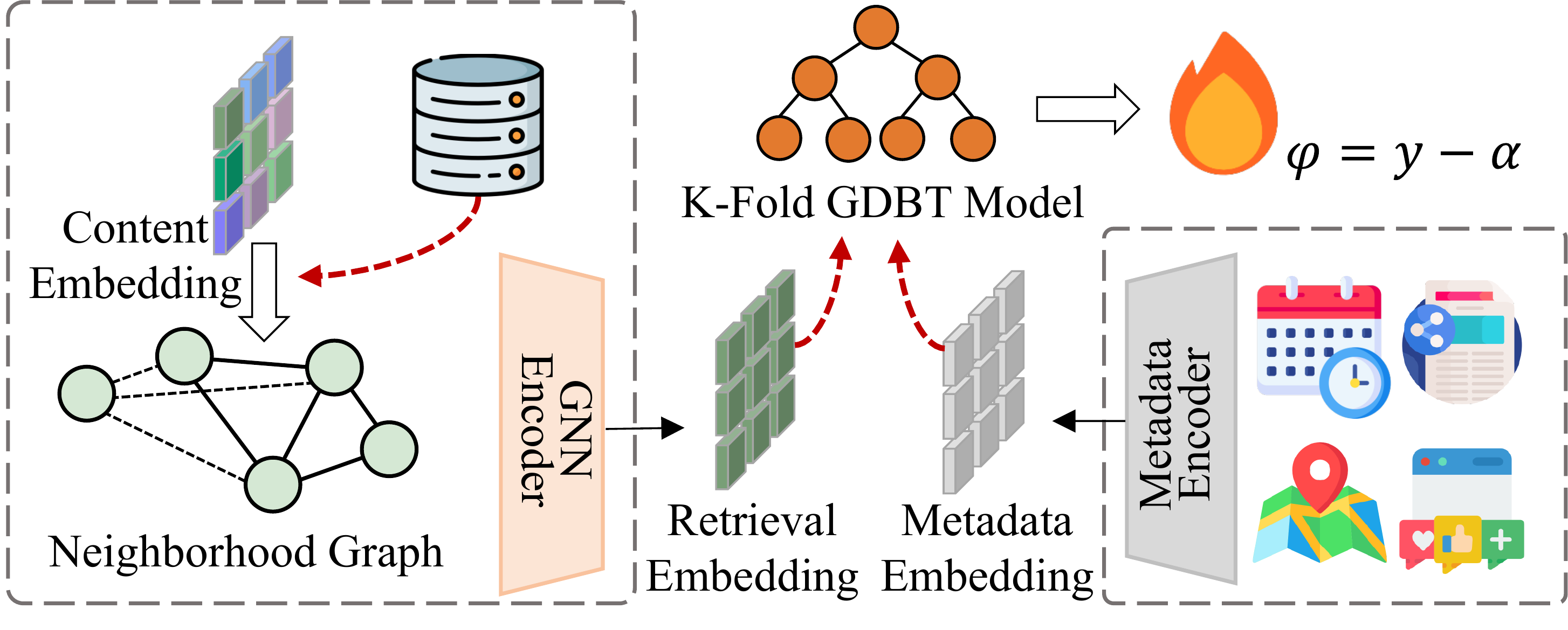}
\caption{
Architecture of the platform-specific context modeling module, which estimates exposure from metadata and retrieved neighborhood cues with a GNN-enhanced predictor.
}
\label{fig:context}
\end{figure}

While attractive content draws attention, a post’s visibility is influenced by external factors that are independent of the content itself.
These factors include temporal rhythms, author activity, topical and regional context, platform delivery mechanisms, and other latent signals whose distributions vary across platforms and over time~\cite{haimovich2022time,cheng2024casdo,jing2026mmcas}.
Modeling these signals separately from content helps isolate platform-driven variability that is not reflected in the post itself.
To model exposure independently from content, the context module estimates an exposure value from exogenous variables and explicitly excludes any content derived embeddings, as shown in \autoref{fig:context}.
The context module alone uses metadata. Retrieval provides aggregated neighborhood statistics rather than raw content embeddings. This design keeps the content module portable across platforms.
The module is platform-specific and captures distribution mechanisms unique to each platform.
For each post $i$, we construct a context vector that concatenates temporal, user, topical, geographic, historical, and platform features:
\begin{equation}
    x_i^{\mathrm{ctx}} = \big[x_i^{\mathrm{time}},\,x_i^{\mathrm{user}},\,x_i^{\mathrm{topic}},\,x_i^{\mathrm{geo}},\,x_i^{\mathrm{hist}},\,x_i^{\mathrm{plat}}\big].
\end{equation}
Temporal features encode cyclic posting patterns using sine and cosine pairs of hour and weekday.
User features summarize follower scale, activity frequency, and recent engagement with counts stabilized by $\log(1+\cdot)$.
Topic and location categories are converted into dense statistics with out of fold target encoding to prevent information leakage.
Historical and platform-specific features describe short and medium horizon dynamics and delivery configurations that affect reach.
All numeric inputs are standardized within each platform to ensure comparable feature and stable optimization.

With the content module frozen, the context module predicts exposure from exogenous inputs as
$\hat{\phi}_i=f_{\phi}\!\big(x_i^{\mathrm{ctx}}\big)$.
Supervision uses the residual target
$r_i = \tilde{y}_i - \hat{\alpha}_i$ with $\tilde{y}_i = \log(1 + y_i)$,
which corresponds to the component of popularity not explained by content attractiveness.
Training minimizes a robust weighted Huber loss with a mild calibration term,
\begin{equation}
\label{eq:loss_phi}
\begin{aligned}
    L_{\phi} = &\frac{1}{B}\sum_{i=1}^{B}w_i\,\mathrm{Huber}_{\delta}(r_i-\hat{\phi}_i) \\
    &+\lambda_{\mathrm{mv}}\Big(|\mathrm{mean}(\hat{\phi})|+|\mathrm{var}(\hat{\phi})-\mathrm{var}(r)|\Big).
\end{aligned}
\end{equation}
To emphasize rare high-exposure samples under a long-tailed label distribution,
the sample weights $w_i$ are assigned by stratifying $\tilde{y}_i$ into three quantile-based bins.
Data are split chronologically so that each prediction depends only on information available in the past.
Each platform maintains its own exposure model $f_{\phi}$, which is trained independently to capture its visibility dynamics.

\subsection{Retrieval-Augmented Neighborhood Module}
\label{sec:neighbor}
Exposure also reflects how collective attention propagates over time.
To capture such diffusion effects without leaking content information into the exposure estimator, we enrich the context with neighborhood aggregates computed only from earlier posts~\cite{zhong2024predicting,cheng2024retrieval}.

Given the content embedding space produced by the content module, we retrieve the top $K$ past neighbors of post $i$ by cosine similarity.
This design follows recent retrieval based popularity models and uses temporal filtering and weighted summary statistics to reduce noise and mismatch with the prediction objective~\cite{zhong2024predicting,cheng2024retrieval,xu2025skapp,xu2026jrpp}.
Each neighbor $j$ receives a weight that combines temporal decay and semantic proximity,
\begin{equation}
\label{eq:w_ij}
\begin{aligned}
    \tilde{w}_{ij} &= \exp\!\Big(-\frac{\Delta t_{ij}}{\tau}\Big)\exp\!\big(s_{ij}-\max_k s_{ik}\big), \\
    w_{ij} &= \frac{\tilde{w}_{ij}}{\sum_{j'}\tilde{w}_{ij'}}.
\end{aligned}
\end{equation}
where $\Delta t_{ij}$ is the temporal gap and $s_{ij}$ is the cosine similarity between post $i$ and its neighbor $j$,
and $\max_k s_{ik}$ denotes the maximum similarity among all retrieved neighbors of post $i$.
Using these weights, we compute a fixed length vector of weighted statistics that summarize recent diffusion activity,
\begin{equation}
\label{eq:nbr}
\begin{aligned}
    x_i^{\mathrm{nbr}} = \Big[&
    \mathbb{E}_w[y_j],\,
    \mathrm{Var}_w[y_j],\,
    \mathbb{E}_w[a_j],\\
    &\mathrm{Var}_w[a_j],\,
    \mathbb{E}_w[\Delta t_{ij}],\,
    \mathrm{Var}_w[\Delta t_{ij}]
    \Big],
\end{aligned}
\end{equation}
where $\mathbb{E}_w[\cdot]$ and $\mathrm{Var}_w[\cdot]$ denote the weighted mean and variance under normalized weights $w_{ij}$, $y_j$ is neighbor popularity, and $a_j$ is neighbor author recent engagement.
By combining contextual and neighborhood features, we obtain the augmented input
$\bar{x}_i = \big[x_i^{\mathrm{ctx}},\,x_i^{\mathrm{nbr}}\big]$, which is mapped to the final exposure estimate
$\hat{\phi}_i = f_{\phi}\!\big(\bar{x}_i\big)$.
This design encodes short-term momentum and social contagion while preserving a clean separation between content signals and exposure modeling, since only aggregated neighborhood statistics rather than raw content embeddings are provided to the exposure estimator.

\section{Experiments}
\label{sec:experiments}

\subsection{Datasets and Evaluation}

We evaluate OmniTrend on four benchmarks covering short-video, image, and cross-platform scenarios.
\textbf{MicroLens}~\cite{ni2023content} contains 1.02\,M short videos with visual, audio, and textual signals.
\textbf{ICIP}~\cite{ortis2019prediction} includes 11.7\,k social images collected from Twitter and Instagram.
\textbf{SMPD-Image} consists of 486\,k images from 70\,k users across 756 categories collected over 16 months with 250\,k tags, and \textbf{SMPD-Video} comprises 6\,k short videos from 4.5\,k users across 120 categories over 24 months with 40\,k tags. These datasets reflect recent multimodal and temporal popularity benchmarks~\cite{wu2024smpchallenge,xu2025smtpd}. All dataset splits follow a chronological 8:1:1 ratio for training, validation, and testing. Retrieval features and neighborhood statistics are computed using only posts that precede the target post in time to ensure causal consistency.
The four benchmarks differ substantially in scale, modality, and popularity regime. This diversity tests whether the content module remains transferable while the context module adapts to platform specific exposure patterns. The supplementary material provides the full dataset overview and label distribution plots.

To ensure a fair and comprehensive evaluation, OmniTrend is compared with a broad range of baseline models widely used for social media popularity prediction, including SVR~\cite{khosla2014makes}, HyFea~\cite{lai2020hyfea}, CLSTM~\cite{ghosh2016contextual}, HMMVED~\cite{xie2021micro}, CBAN~\cite{cheung2022crossmodal}, BLIP~\cite{li2022blip}, MASSL~\cite{zhang2022multi}, NIPA~\cite{ji2023community}, MMRA~\cite{zhong2024predicting}, TMALL~\cite{chen2016micro}, and DLBA~\cite{viola2021instagram}. These baselines cover feature based, sequential, multimodal, graph, and retrieval based methods. All models are trained and evaluated on the same chronological splits using the same label transformation $\log(1+y)$ for consistent comparison.

For performance measurement, we report mean squared error (MSE), mean absolute error (MAE), and Spearman rank correlation (SRC). Lower MSE and MAE indicate smaller prediction errors, while higher SRC reflects stronger rank consistency between predictions and ground truth labels. Unless otherwise specified, results are averaged over three random seeds with identical chronological data partitions.

\subsection{Implementation Details}

The content module is trained for five epochs using AdamW with a learning rate of $2\times10^{-4}$, weight decay of $10^{-2}$, a cosine decay schedule, and a batch size of 256. 
Training combines a weighted Huber loss with $\delta=1.0$, a pairwise margin ranking term, and a cross-modal alignment loss, using sample weights of 1.0, 1.5, and 3.0 for the low-, medium-, and high-popularity bins. 
The context module is optimized on residual popularity with CatBoost~\cite{prokhorenkova2018catboost} using Huber loss, a learning rate of 0.03, tree depth of 8, and early stopping after 200 rounds. 
It models temporal, author, and topic factors together with neighbor features obtained from the ten most similar past posts, weighted by an exponential time decay of $\tau=86{,}400$ seconds. 
All datasets are split chronologically in an 8:1:1 ratio for training, validation, and testing, and results are averaged over three random seeds.

To ensure consistent and reproducible context modeling, we implement the temporal, user, topic, and geographic features using simple and standardized transformations. 
Temporal features encode posting hour and weekday through sine and cosine pairs. 
User features include log-transformed follower count, posting frequency, and recent engagement statistics. 
Topic and location identifiers are converted into dense numerical features through out-of-fold target encoding to prevent label leakage. 
For the retrieval module, we build a content embedding index for each platform and restrict neighbor search to posts that occur strictly earlier than the target post.
The weighted statistics in \autoref{eq:nbr} are then computed from the ten retrieved neighbors and concatenated with the contextual features to serve as inputs for exposure prediction.
The supplementary material further shows stable optimization and steady validation improvement for the content module under this training setup.

\begin{figure}[!t]
\centering
\includegraphics[width=\linewidth]{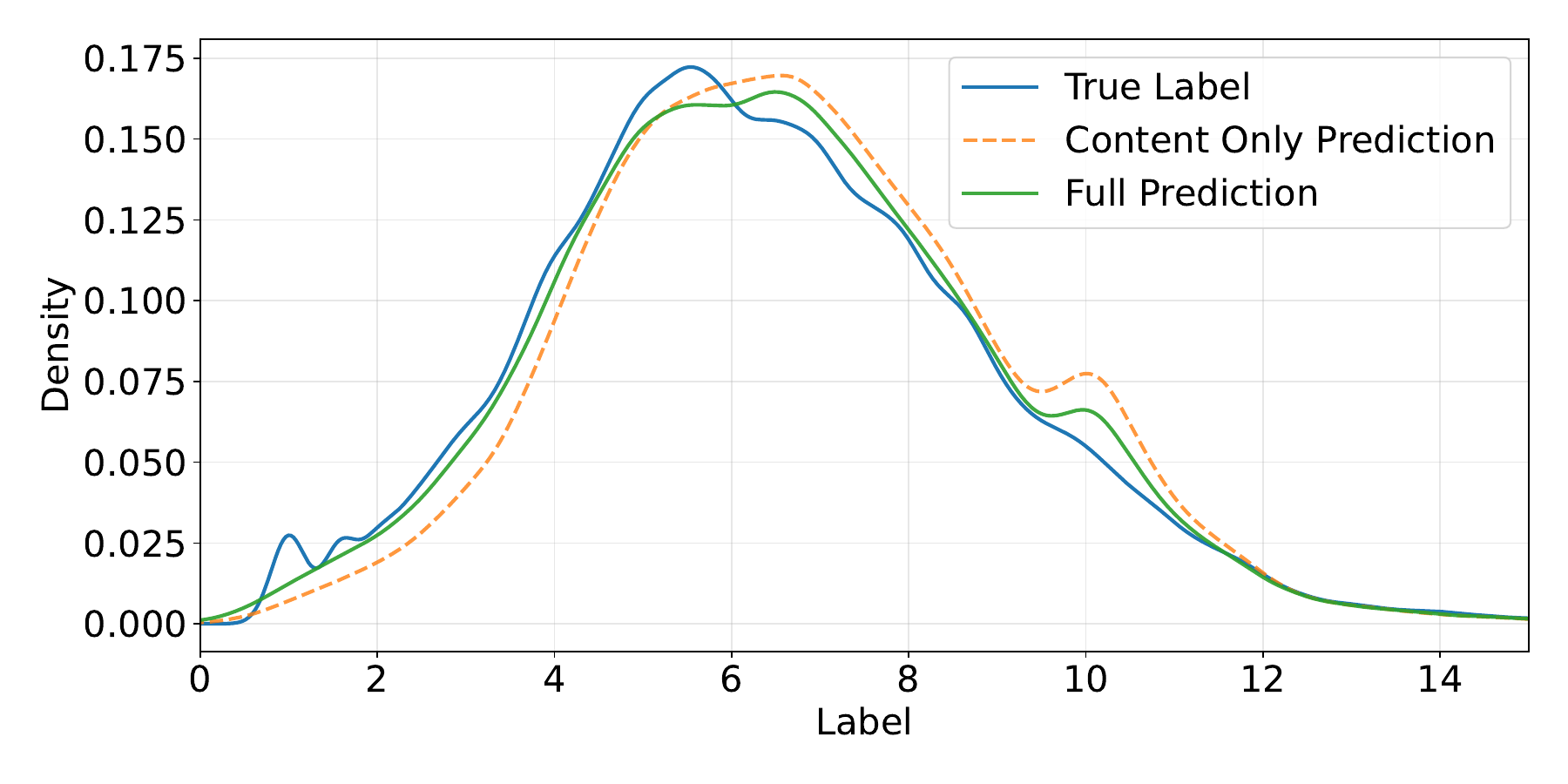}
\caption{
Distribution comparison of true labels, content-only prediction, and full prediction.
}
\label{fig:label_distribution}
\end{figure}

\subsection{Main Results}

\paragraph{Image-Centric Social Media Popularity Prediction.}

\begin{table}
\centering
\caption{
Image-centric comparison on SMPD-Image and ICIP datasets. 
The highest-ranking are in \textbf{bold}; The second-highest is in \underline{underlined}.
}
\label{tab:main_image}
\resizebox{\linewidth}{!}{
\begin{tabular}{lcccccc}
\toprule
\multirow{2}{*}{Method} & \multicolumn{3}{c}{SMPD-Image} & \multicolumn{3}{c}{ICIP} \\
\cmidrule(lr){2-4} \cmidrule(lr){5-7}
& MSE$\downarrow$ & MAE$\downarrow$ & SRC$\uparrow$ & MSE$\downarrow$ & MAE$\downarrow$ & SRC$\uparrow$ \\
\midrule
SVR & 8.13 & 2.41 & 0.51 & 1.90 & 0.89 & 0.52 \\
HyFea & 8.11 & 2.37 & 0.56 & 1.90 & 1.02 & 0.45 \\
MFTM & 4.02 & 1.55 & 0.58 & 1.90 & 0.98 & 0.42 \\
CLSTM & 3.91 & 1.50 & 0.59 & 1.87 & 0.98 & 0.47 \\
HMMVED & 3.72 & \underline{1.36} & 0.64 & 1.86 & 0.95 & 0.45 \\
DLBA & 4.87 & 1.70 & 0.44 & 2.23 & 1.01 & 0.36 \\
MASSL & 5.57 & 1.84 & 0.53 & 1.94 & 0.93 & 0.45 \\
BLIP & 4.39 & 1.63 & 0.53 & 2.06 & 1.00 & 0.36 \\
CBAN & 4.04 & 1.51 & 0.58 & 1.81 & 0.93 & 0.47 \\
NIPA & 4.25 & 1.65 & 0.41 & 2.00 & 1.00 & 0.40 \\
MMRA & \underline{3.51} & 1.37 & \underline{0.64} & \underline{1.76} & \underline{0.87} & \underline{0.54} \\
\textbf{OmniTrend} & \textbf{3.06} & \textbf{1.31} & \textbf{0.71} & \textbf{1.70} & \textbf{0.66} & \textbf{0.89} \\
\bottomrule
\end{tabular}
}
\end{table}

\autoref{tab:main_image} reports the performance of OmniTrend and nine baseline methods on two real-world image datasets, \textbf{SMPD-Image} and \textbf{ICIP}.  
Traditional feature-based models such as SVR, HyFea, and MFTM achieve moderate accuracy on smaller datasets but degrade significantly on larger benchmark.  
Deep multimodal models (CLSTM, HMMVED, CBAN, BLIP) show improved performance, yet their SRC values remain below 0.65, revealing limited ability to capture fine-grained visual–textual interactions.  
Retrieval-augmented models like MMRA outperform earlier deep architectures by leveraging image–text alignment, yet they still fall short of our unified approach.

\textbf{OmniTrend} achieves the best results across all metrics.
On SMPD-Image, it reduces MSE from $3.51$ to $3.06$ and raises SRC from $0.64$ to $0.71$ compared with MMRA, while on ICIP it lowers MSE from $1.76$ to $1.70$ and increases SRC from $0.54$ to $0.89$.
These consistent gains demonstrate that jointly modeling multimodal attractiveness and contextual exposure yields stronger generalization across datasets of different scale and platform diversity.

\paragraph{Video-Centric Social Media Popularity Prediction.}
\autoref{tab:main_video} summarizes results on two large-scale video benchmarks, \textbf{MicroLens} and \textbf{SMPD-Video}.  
Classical regression and handcrafted feature models (SVR, HyFea) show limited ranking consistency, while sequence-based CLSTM and HMMVED struggle with long-term temporal dynamics.  
Retrieval-based MMRA and fusion-based MFTM improve correlation, but they remain sensitive to noisy frame sampling, as their overall performance depends on the stability of frame-level features during retrieval and fusion.

\begin{table}[!t]
\centering
\setlength{\tabcolsep}{3pt}
\renewcommand{\arraystretch}{0.9}
\caption{
Video-centric comparison on MicroLens and SMPD-Video datasets.
The highest-ranking are in \textbf{bold}; The second-highest is in \underline{underlined}.
}
\label{tab:main_video}
\resizebox{\linewidth}{!}{
\begin{tabular}{lcccccc}
\toprule
\multirow{2}{*}{Method} & \multicolumn{3}{c}{MicroLens} & \multicolumn{3}{c}{SMPD-Video} \\
\cmidrule(lr){2-4} \cmidrule(lr){5-7}
& MSE$\downarrow$ & MAE$\downarrow$ & SRC$\uparrow$ & MSE$\downarrow$ & MAE$\downarrow$ & SRC$\uparrow$ \\
\midrule
SVR & 1.58 & 1.22 & 0.43 & 4.92 & 1.80 & 0.48 \\
HyFea & 1.56 & 1.23 & 0.43 & \underline{4.01} & \underline{1.62} & 0.57 \\
MFTM & 1.73 & 1.70 & 0.31 & 4.57 & 1.75 & \underline{0.66} \\
CLSTM & 1.54 & 1.21 & 0.46 & 5.71 & 1.97 & 0.57 \\
HMMVED & 1.66 & 1.25 & 0.37 & 5.58 & 1.83 & 0.43 \\
MASSL & 2.08 & 1.41 & 0.39 & 4.75 & 1.75 & 0.52 \\
CBAN & 1.49 & 1.19 & 0.47 & 4.35 & 1.67 & 0.55 \\
TMALL & 1.68 & 1.32 & 0.34 & 7.30 & 2.15 & 0.51 \\
MMRA & \underline{1.45} & \underline{1.18} & \underline{0.49} & 5.12 & 1.84 & 0.50 \\
\textbf{OmniTrend} & \textbf{1.07} & \textbf{0.86} & \textbf{0.61} & \textbf{3.92} & \textbf{1.44} & \textbf{0.66} \\
\bottomrule
\end{tabular}
}
\end{table}

In contrast, \textbf{OmniTrend} achieves the lowest errors and the highest correlation on both datasets.  
On MicroLens, it reduces MAE from $1.18$ to $0.86$ and raises SRC from $0.49$ to $0.61$.  
On SMPD-Video, it further reduces MAE from $1.62$ to $1.44$ and improves SRC from $0.655$ to $0.662$, surpassing all baselines by a clear margin.  
Performance improvements are especially evident in long-tail popularity bins, where relational aggregation compensates for missing textual cues and stabilizes prediction under sparse contexts.

\subsection{Ablation Studies}
Unless mentioned otherwise, we report ablation results on SMPD-Image dataset.

\paragraph{Core Module Analysis.}
\autoref{tab:ablation_core} analyzes the contribution of the Content and Context modules.
Removing the Context module causes a substantial increase in both MSE and MAE, indicating that exposure cues play a key role in stabilizing predictions.
Removing the Content module also degrades performance, especially in correlation, showing that intrinsic attractiveness remains indispensable.
The complete model, which combines the two modules, achieves the best accuracy and ranking consistency.

\begin{figure}[t]
    \centering
    \includegraphics[width=0.85\linewidth]{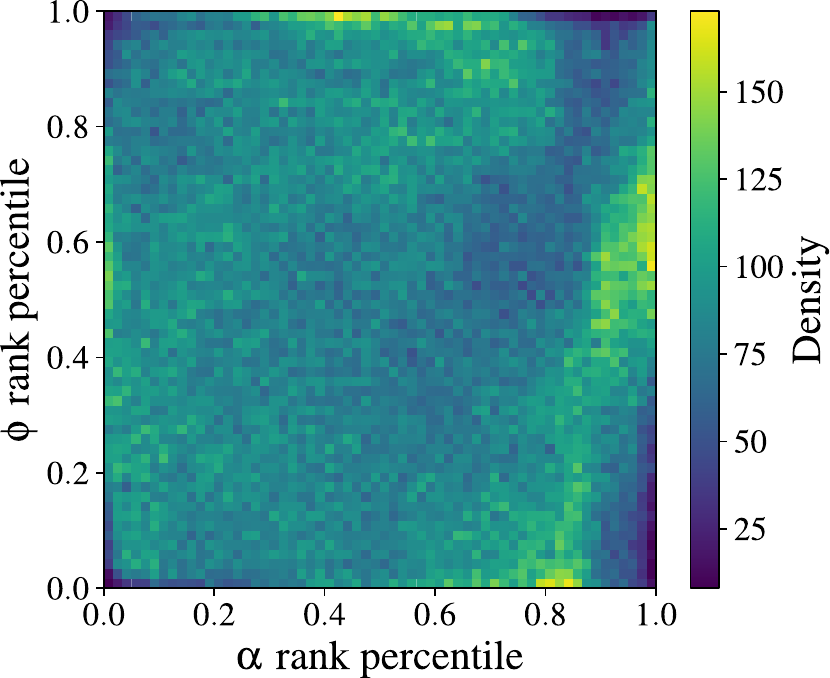}
    \caption{
        Rank--rank heatmap between the predicted content score $\hat{\alpha}$ and context score $\hat{\phi}$ on the test set.
        The uniform density distribution indicates that the two predicted factors are statistically independent.
    }
    \label{fig:rankrank}
\end{figure}

To further examine whether the two modules capture distinct factors, we visualize the joint distribution of the predicted content score $\hat{\alpha}$ and context score $\hat{\phi}$ on the test set, as shown in \autoref{fig:rankrank}. The nearly uniform rank–rank distribution suggests that the two scores are statistically independent, confirming that the model effectively disentangles content attractiveness from contextual exposure.

To further illustrate the effectiveness of context modeling, we compare the distribution of predicted popularity scores with the ground truth labels.
As shown in \autoref{fig:label_distribution}, the full model with both content and contextual information produces a distribution that closely resembles the true label distribution, demonstrating its effectiveness in capturing real-world popularity dynamics.
These results confirm that joint content and context modeling is essential for robust popularity estimation.
The supplementary material also provides a direct diagnostic of the additive decomposition. The content score follows the broad trend of the target distribution, while the combined score is much closer to the identity relation with the ground truth labels. This behavior is consistent with the residual exposure formulation used in the model.

\begin{table}
\centering
\caption{Ablation of Content and Context modules.
We examine the individual and combined effects of the Content and Context modules on SMPD-Image.}
\label{tab:ablation_core}
\begin{tabular}{lccc}
\toprule
Variant & MSE$\downarrow$ & MAE$\downarrow$ & SRC$\uparrow$ \\
\midrule
w/o Context module & 4.06 & 2.62 & 0.52 \\
w/o Content module & 4.07 & 1.59 & 0.65 \\
\textbf{Full model} & \textbf{3.06} & \textbf{1.31} & \textbf{0.71} \\
\bottomrule
\end{tabular}
\end{table}

\paragraph{Cross platform training of the Content module.}
\autoref{tab:abl_cross_alpha} compares different training sources for the Content module while keeping the Context module on SMPD-Image.
Training on SMPD-Video alone performs worst.
Using SMPD-Image alone improves accuracy and correlation.
Joint training on SMPD-Image and SMPD-Video matches the full OmniTrend configuration, confirming that cross-platform content learning is already adopted in our main model.

\begin{table}
\centering
\caption{Cross platform training of the Content module evaluated on SMPD-Image.
$\alpha$ denotes the Content module and $\phi$ denotes the Context module.}
\label{tab:abl_cross_alpha}
\begin{tabular}{lccc}
\toprule
Configuration & MSE$\downarrow$ & MAE$\downarrow$ & SRC$\uparrow$ \\
\midrule
$\alpha$: Video & 3.84 & 1.63 & 0.67 \\
$\alpha$: Image & 3.21 & 1.45 & 0.69 \\
\textbf{$\alpha$: Image + Video} & \textbf{3.06} & \textbf{1.31} & \textbf{0.71} \\
\bottomrule
\end{tabular}
\end{table}

\begin{table}[!htbp]
\centering
\caption{Ablation of key components in the Context module on SMPD-Image.}
\label{tab:abl_context}
\begin{tabular}{lccc}
\toprule
Variant & MSE$\downarrow$ & MAE$\downarrow$ & SRC$\uparrow$ \\
\midrule
w/o retrieval & 4.06 & 1.59 & 0.64 \\
w/o contextual features & 4.63 & 1.70 & 0.60 \\
\textbf{Full Context Module} & \textbf{3.06} & \textbf{1.31} & \textbf{0.71} \\
\bottomrule
\end{tabular}
\end{table}

\paragraph{Transferability under limited context.}
Beyond the standard ablations, the learned attractiveness signal remains informative when contextual metadata are sparse. On a cold start subset of SMPD-Image, the content only branch attains an MSE of $2.35$, an MAE of $1.55$, and an SRC of $0.485$. The full model improves these results to $2.18$, $1.49$, and $0.532$ when contextual and historical cues are available. In transfer from images to videos, a frozen content branch trained on images reaches an SRC of $0.602$, compared with $0.662$ under supervised training on videos. These results indicate that the content module captures attractiveness cues that transfer across image and video domains.

\paragraph{Context Module Analysis.}
\autoref{tab:abl_context} analyzes the key components in the Context module.
Removing the retrieval aggregation significantly increases error and reduces correlation, indicating that neighborhood information is essential for capturing temporal diffusion and social contagion effects.
Eliminating contextual features such as temporal and user attributes further degrades performance, showing that exposure cannot be accurately estimated from platform-independent statistics alone.
By integrating temporal cues and retrieval-based aggregation, the complete context module attains the highest accuracy and ranking consistency across all evaluated settings.
The supplementary material further reports feature importance for the context model. User activity statistics, semantic category cues, and neighborhood aggregates are among the strongest exposure signals. This pattern is consistent with the ablation trend, and the full ranking appears in the supplementary material.

\begin{table}[!htbp]
\centering
\caption{Modal contributions in the Content module.
We remove either the visual or textual input to assess modality importance on SMPD-Image.}
\label{tab:abl_modal_alpha}
\begin{tabular}{lccc}
\toprule
Variant & MSE$\downarrow$ & MAE$\downarrow$ & SRC$\uparrow$ \\
\midrule
Without visual input & 3.56 & 1.70 & 0.67 \\
Without textual input & 3.42 & 1.59 & 0.68 \\
\textbf{Image + Text} & \textbf{3.06} & \textbf{1.31} & \textbf{0.71} \\
\bottomrule
\end{tabular}
\end{table}

\paragraph{Modal Contributions in the Content Module.}
\autoref{tab:abl_modal_alpha} evaluates the importance of visual and textual modalities in the Content module.
Removing the visual input leads to higher MSE and MAE and a notable drop in SRC, suggesting that image cues are crucial for estimating attractiveness.
Removing the textual input causes a smaller yet consistent degradation, indicating that language features complement visual perception.
When both modalities are used together, the model achieves the lowest prediction errors and the highest rank correlation, which demonstrates that multimodal content understanding is essential for accurate popularity prediction.

\subsection{Summary of Findings}
OmniTrend consistently outperforms all baselines across four benchmarks.
The joint modeling of content and context improves both numerical accuracy and rank consistency while maintaining clear interpretability.
Ablation studies confirm that each module contributes distinct and complementary information:
the Content module captures multimodal attractiveness that transfers across platforms, and the Context module estimates exposure dynamics through temporal and neighborhood cues.
Retrieval-enhanced aggregation further stabilizes predictions in long-tail and rapidly changing scenarios.
This advantage remains consistent across both image and video settings, despite large differences in scale, modality, and temporal dynamics.
Overall, OmniTrend provides an accurate, transferable, and interpretable framework for social media popularity prediction across diverse modalities and platforms.
These observations indicate that the explicit separation between attractiveness and exposure not only strengthens predictive performance but also offers a reliable basis for examining how popularity evolves under different social platform and temporal conditions.

Beyond these empirical results, the modular separation between content attractiveness and contextual exposure also benefits practical adaptation.
The content module can be upgraded with stronger multimodal encoders as they become available, while the context module can be updated to incorporate newly observed temporal or author-level signals without retraining the entire framework.
This design allows OmniTrend to track platform evolution at different timescales while preserving a stable notion of content attractiveness under temporal shifts.

\section{Conclusion and Future Work}

This work presents \textbf{OmniTrend}, a unified framework for social media popularity prediction that jointly models \textit{content attractiveness} and \textit{contextual exposure}.
By separating these factors into complementary modules, OmniTrend provides interpretable and transferable predictions across platforms.
The content module captures intrinsic multimodal appeal, while the context module estimates exposure dynamics with retrieval-enhanced temporal cues.

Experiments on multiple image and video benchmarks show that OmniTrend consistently outperforms existing multimodal and retrieval-based methods in both accuracy and ranking stability.
These results further suggest that separating attractiveness from exposure helps maintain robustness across platforms with different modalities, scales, and temporal patterns.
Future extensions include incorporating fine-grained user feedback for exposure modeling,
adopting large multimodal backbones for richer content understanding,
and broadening transfer evaluation to more platforms and adaptation settings.
Another direction is to study exposure aware evaluation across time horizons and improve retrieval robustness under temporal drift~\cite{haimovich2022time,xu2025smtpd,xu2025skapp,xu2026jrpp}.
Evaluating OmniTrend on a broader range of image and video platforms would provide a more complete view of its behavior across platforms in practical settings.
Additional transfer evaluations could further test how the framework responds to changes in platform design, content formats, and user interaction patterns~\cite{liu2025popsim}.

\bibliographystyle{ACM-Reference-Format}
\bibliography{main}


\end{document}